\documentclass[11pt,a4paper]{article}
\usepackage[hyperref]{acl2018}
\usepackage{times}
\usepackage{latexsym}

\usepackage{url}

\usepackage{graphicx}
\usepackage{booktabs}
\usepackage{tabularx}
\usepackage{amsmath}
\usepackage{microtype}

\usepackage{csquotes}
\usepackage{pgfplots}

\aclfinalcopy 

\title{Scoring Lexical Entailment \\ with a Supervised Directional Similarity Network}

\author{Marek Rei$^{\spadesuit\diamondsuit}$ ~ Daniela Gerz$^{\clubsuit}$ ~ Ivan Vuli\'c$^{\clubsuit}$ \\
$^\spadesuit$Computer Laboratory, University of Cambridge, United Kingdom \\
$^\diamondsuit$The ALTA Institute, University of Cambridge, United Kingdom \\
$^\clubsuit$Language Technology Lab, University of Cambridge, United Kingdom \\
{ \small \tt
marek.rei@cl.cam.ac.uk, dsg40@cam.ac.uk, iv250@cam.ac.uk
}
}

\date{}

\begin{document}
\maketitle
\begin{abstract}

We present the Supervised Directional Similarity Network (SDSN), a novel neural architecture for learning task-specific transformation functions on top of general-purpose word embeddings. 
Relying on only a limited amount of supervision from task-specific scores on a subset of the vocabulary, our architecture is able to generalise and transform a general-purpose distributional vector space to model the relation of lexical entailment. Experiments show excellent performance on scoring graded lexical entailment, raising the state-of-the-art on the HyperLex dataset by approximately 25\%. 
\end{abstract}

\section{Introduction}
Standard word embedding models \cite{Mikolova,Pennington:2014emnlp,Bojanowski:2017tacl} are based on the distributional hypothesis by \newcite{Harris:1954}.  However, purely distributional models coalesce various lexico-semantic relations (e.g., synonymy, antonymy, hypernymy) into a joint distributed representation. To address this, previous work has focused on introducing supervision into \textit{individual} word embeddings, allowing them to better capture the desired lexical properties. For example, \newcite{Faruqui2015} and \newcite{Wieting:2015tacl} proposed methods for using annotated lexical relations to condition the vector space and bring synonymous words closer together. \newcite{Mrksic2016} and \newcite{Mrksic2017} improved the optimisation function and introduced an additional constraint for pushing antonym pairs further apart. While these methods integrate hand-crafted features from external lexical resources with distributional information, they improve only the embeddings of words that have annotated lexical relations in the training resource.

In this work, we propose a novel approach to leveraging external knowledge with general-purpose unsupervised embeddings, focusing on the directional graded lexical entailment task \cite{Vulic2017}, whereas previous work has mostly investigated simpler non-directional semantic similarity tasks. Instead of optimising individual word embeddings, our model uses general-purpose embeddings and optimises a separate neural component to adapt these to the specific task. In particular, our neural Supervised Directional Similarity Network (SDSN) dynamically produces task-specific embeddings optimised for scoring the asymmetric lexical entailment relation between any two words, regardless of their presence in the training resource. Our results with task-specific embeddings indicate large improvements on the HyperLex dataset, a standard graded lexical entailment benchmark. The model also yields improvements on a simpler non-graded entailment detection task.





\section{The Task of Grading Lexical Entailment}
In graded lexical entailment, the goal is to make fine-grained assertions regarding the directional hierarchical semantic relationships between concepts \cite{Vulic2017}. The task is grounded in theories of concept (proto)typicality and category vagueness from cognitive science \cite{Rosch:1975cognitive,Kamp:1995cog}, and aims at answering the following question: \textit{``To what degree is $X$ a type of $Y$?''}. It quantifies the degree of lexical entailment instead of providing only a binary \textit{yes/no} decision on the relationship between the concepts $X$ and $Y$, as done in hypernymy detection tasks 
\cite{Kotlerman:2010nle,Weeds:2014coling,Santus:2014eacl,Kiela:2015acl,Shwartz:2017eacl}.

Graded lexical entailment provides finer-grained judgements on a continuous scale. For instance, the word pair (\textit{girl $\rightarrow$ person}) has been rated highly with 9.85/10 by the HyperLex annotators.
The pair (\textit{guest $\rightarrow$ person}) has received a slightly lower score of $7.22$, as a prototypical guest is often a person but there can be exceptions.
In contrast, the score for the reversed pair (\textit{person $\rightarrow$ guest}) is only judged at $2.88$.

As demonstrated by \newcite{Vulic2017} and \newcite{Nickel2017}, standard general-purpose representation models trained in an unsupervised way purely on distributional information are unfit for this task and unable to surpass the performance of simple frequency baselines (see also Table~\ref{tab:results}). Therefore, in what follows, we describe a novel supervised framework for constructing task-specific word embeddings, optimised for the graded entailment task at hand. 

\section{System Architecture}
\label{ss:architecture}
\begin{figure}[t]
	\centering
	\includegraphics[width=0.9\linewidth]{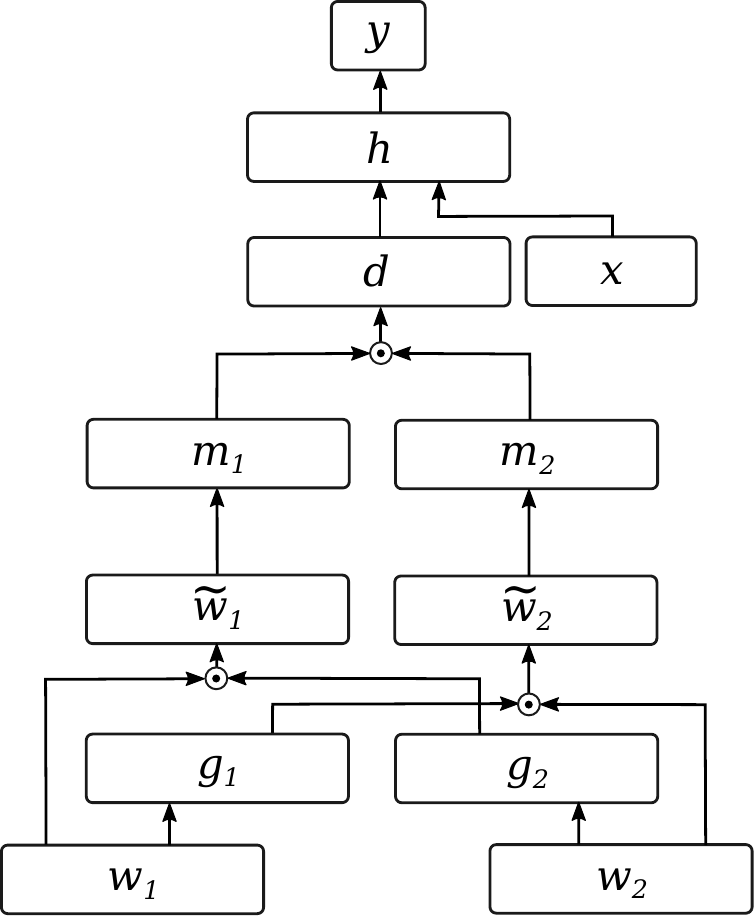}
	\caption{Supervised directional similarity network (SDSN) for grading lexical relations.}
	\label{fig:network}
\end{figure}

The network architecture can be seen in Figure~\ref{fig:network}. The system receives a pair of words as input and predicts a score that represents the strength of the given lexical relation. 
In the graded entailment task, we would like the model to return a high score for \textit{(biology $\rightarrow$ science)}, as biology is a type of science, but a low score for \textit{(point $\rightarrow$ pencil)}.

We start by mapping both input words to corresponding word embeddings $w_1$ and $w_2$. The embeddings come from a standard distributional vector space, pre-trained on a large unannotated corpus, and are not fine-tuned during training. 
An element-wise gating operation is then applied to each word, conditioned on the other word:
\vspace{-0mm}
\begin{align}
&g_1 = \sigma(W_{g_1} w_1 + b_{g_1}) \\
&g_2 = \sigma(W_{g_2} w_2 + b_{g_2}) \\
&\widetilde{w}_1 = w_1 \odot g_2 \\
&\widetilde{w}_2 = w_2 \odot g_1
\end{align}

\noindent where $W_{g_1}$ and $W_{g_2}$ are weight matrices, $b_{g_1}$ and $b_{g_2}$ are bias vectors, $\sigma()$ is the logistic function and $\odot$ indicates element-wise multiplication. This operation allows the network to first observe the candidate hypernym $w_2$ and then decide which features are important when analysing the hyponym $w_1$. For example, when deciding whether \textit{seal} is a type of \textit{animal}, the model is able to first see the word \textit{animal} and then apply a mask that blocks out features of the word \textit{seal} that are not related to nature. 
During development we found it best to apply this gating in both directions, therefore we condition each word based on the other.

Each of the word representations is then passed through a non-linear layer with $tanh$ activation, mapping the words to a new space that is more suitable for the given task:
\begin{align}
&m_1 = tanh(W_{m_1} \widetilde{w}_1 + b_{m_1}) \\
&m_2 = tanh(W_{m_2} \widetilde{w}_2 + b_{m_2})
\end{align}

\noindent where $W_{m_1}$, $W_{m_2}$, $b_{m_1}$ and $b_{m_2}$ are trainable parameters.
The input embeddings are trained to predict surrounding words on a large unannotated corpus using the skip-gram objective \cite{Mikolova}, making the resulting vector space reflect (a broad relation of) semantic relatedness but unsuitable for lexical entailment \cite{Vulic2017}. The mapping stage allows the network to learn a transformation function from the general skip-gram embeddings to a task-specific space for lexical entailment.
In addition, the two weight matrices enable asymmetric reasoning, allowing the network to learn separate mappings for hyponyms and hypernyms. 

We then use a supervised composition function for combining the two representations and returning a confidence score as output.
\newcite{Rei2017a} described a generalised version of cosine similarity for metaphor detection, constructing a supervised operation and learning individual weights for each feature. We apply a similar approach here and modify it to predict a relation score:
\begin{align}
&d = m_1 \odot m_2\\
&h = tanh(W_h d + b_h) \\
&y = S \cdot \sigma(a(W_y h + b_y))
\end{align}

\noindent where $W_h$, $b_h$, $a$, $W_y$ and $b_y$ are trainable parameters. 
The annotated labels of lexical relations are generally in a fixed range, therefore we base the output function on logistic regression, which also restricts the range of the predicted scores. $b_y$ allows for the function to be shifted as necessary and $a$ controls the slope of the sigmoid. $S$ is the value of the maximum score in the dataset, scaling the resulting value to the correct range.
The output $y$ represents the confidence that the two input words are in a lexical entailment relation.

We optimise the model by minimising the mean squared distance between the predicted score $y$ and the gold-standard score $\hat{y}$:

\begin{equation}
L = \sum_i (y_i - \hat{y_i})^2
\end{equation}

\paragraph{Sparse Distributional Features (SDF).}

Word embeddings are well-suited for capturing distributional similarity, but they have trouble encoding features such as word frequency, or the number of unique contexts the word has appeared in. 
This information becomes important when deciding whether one word entails another, as the system needs to determine when a concept is more general and subsumes the other.

We construct classical sparse distributional word vectors and use them to extract 5 unique features for every word pair, to complement the features extracted from neural embeddings:

\begin{itemize}
\item Regular cosine similarity between the sparse distributional vectors of both words.
\item The sparse weighted cosine measure, described by \newcite{Rei2014}, comparing the weighted ranks of different distributional contexts. The measure is directional and assigns more importance to the features of the broader term. We include this weighted cosine in both directions.
\item The proportion of shared unique contexts, compared to the number of contexts for one word. This measure is able to capture whether one of the words appears in a subset of the contexts, compared to the other word. This feature is also directional and is therefore included in both directions.
\end{itemize}

We build the sparse distributional word vectors from two versions of the British National Corpus \cite{leech1992100}. The first counts contexts simply based on a window of size 3. The second uses a parsed version of the BNC \cite{Andersen2008} and extracts contexts based on dependency relations. In both cases, the features are weighted using pointwise mutual information.
Each of the five features is calculated separately for the two vector spaces, resulting in 10 corpus-based features.
We integrate them into the network by conditioning the hidden layer $h$ on this vector:

\begin{equation}
h = tanh(W_h d + W_x x + b_h)
\end{equation}

\noindent where $x$ is the feature vector of length 10 and $W_x$ is the corresponding weight matrix.


\paragraph{Additional Supervision (AS).}
Methods such as retrofitting \cite{Faruqui2015}, \textsc{attract-repel} \cite{Mrksic2017} and Poincar\'e embeddings \cite{Nickel2017} make use of hand-annotated lexical relations for optimising word representations such that they capture the desired properties (so-called \textit{embedding specialisation}).
We also experiment with incorporating these resources, but instead of adjusting the individual word embeddings, we use them to optimise the shared network weights.
This teaches the model to find useful regularities in general-purpose word embeddings, which can then be equally applied to all words in the embedding vocabulary.

For hyponym detection, we extract examples from WordNet \cite{Miller1995} and the Paraphrase Database (PPDB 2.0) \cite{Pavlick2015}.
We use WordNet synonyms and hyponyms as positive examples, along with antonyms and hypernyms as negative examples. In order to prevent the network from biasing towards specific words that have numerous annotated relations, we limit them to a maximum of 10 examples per word.
From the PPDB we extract the Equivalence relations as positive examples and the Exclusion relations as negative word pairs.

The final dataset contains 102,586 positive pairs and 42,958 negative pairs. 
However, only binary labels are attached to all word pairs, whereas the task requires predicting a graded score.
Initial experiments with optimising the network to predict the minimal and maximal possible score for these cases did not lead to improved performance.
Therefore, we instead make use of a hinge loss function that optimises the network to only push these examples to the correct side of the decision boundary:

\begin{equation}
L = \sum_i max((y-\hat{y})^2 - (\frac{S}{2} - R)^2, 0)
\label{eq:hinge}
\end{equation}
\noindent where $S$ is the maximum score in the range and and $R$ is a margin parameter. 
By minimising Equation~\ref{eq:hinge}, the model is only updated based on examples that are not yet on the correct side of the boundary, including a margin. This prevents us from penalising the model for predicting a score with slight variations, as the extracted examples are not annotated with sufficient granularity.
When optimising the model, we first perform one pre-training pass over these additional word pairs before proceeding with the regular training process.

\section{Evaluation}

\paragraph{SDSN Training Setup.}
As input to the {\small{SDSN}} network we use 300-dimensional dependency-based word embeddings by \newcite{Levy2014a}.
Layers $m_1$ and $m_2$ also have size 300 and layer $h$ has size 100.
For regularisation, we apply dropout to the embeddings with $p=0.5$.
The margin $R$ is set to $1$ for the supervised pre-training stage.
The model is optimised using AdaDelta \cite{Zeiler2012} with learning rate $1.0$.
In order to control for random noise, we run each experiment with 10 different random seeds and average the results. 
Our code and detailed configuration files will be made available online.\footnote{http://www.marekrei.com/projects/sdsn}

\paragraph{Evaluation Data.}
We evaluate graded lexical entailment on the HyperLex dataset \cite{Vulic2017} which contains 2,616 word pairs in total scored for the asymmetric graded lexical entailment relation. Following a standard practice, we report Spearman's $\rho$ correlation of the model output to the given human-annotated scores. We conduct experiments on two standard data splits for supervised learning: \textit{random split} and \textit{lexical} split. In the random split the data is randomly divided into training, validation, and test subsets containing 1831, 130, and 655 word pairs, respectively. In the \textit{lexical split}, proposed by \newcite{Levy:2015naacl}, there is no lexical overlap between training and test subsets. This prevents the effect of \textit{lexical memorisation}, as supervised models tend to learn an independent property of a single concept in the pair instead of learning a relation between the two concepts. In this setup training, validation, and test sets contain 1133, 85, and 269 word pairs, respectively.\footnote{Note that the lexical split discards all cross-set training-test word pairs. Consequently, the 
number of instances in each subset is lower than with the random split.}

Since plenty of related research on lexical entailment is still focused on the simpler binary detection of asymmetric relations, we also run experiments on the large binary detection HypeNet dataset \cite{Shwartz2016}, where the {\small{SDSN}} output is converted to binary decisions. We again report scores for both random and lexical split.

\begin{table}[t]
\centering
\small
\def\arraystretch{1.0}
\setlength\tabcolsep{5pt}
\begin{tabular}{l|cc|cc} \toprule
 & \multicolumn{2}{c|}{Random} & \multicolumn{2}{c}{Lexical} \\
 & \small{DEV} &  \small{TEST} &\small{DEV} &  \small{TEST}\\ \midrule
FR & - & 0.299 & - & 0.199\\
SGNS-DEPS & - & 0.250 & - & 0.253\\ \midrule
WN-WuP & - & 0.212 & - & 0.261\\
SGNS-DEPS (concat+r) & - & 0.539 & - & 0.399\\
Paragram+CF (cos) & - & 0.346 & - & 0.453\\
Paragram+CF (mul+r) & - & 0.386 & - & 0.439\\ \midrule
SDSN & 0.708 & 0.658 & 0.547 & 0.475\\
SDSN+SDF & 0.722 & 0.671 & 0.562 & 0.495\\
SDSN+SDF+AS & \textbf{0.757} & \textbf{0.692} & \textbf{0.577} & \textbf{0.544} \\ \bottomrule
\end{tabular}
\caption{Graded lexical entailment detection results on the random and lexical splits of the HyperLex dataset. We report Spearman's $\rho$ on both validation and test sets. }
\label{tab:results}
\end{table}

\paragraph{Results and Analysis.}
The results on two HyperLex splits are presented in Table \ref{tab:results}, along with the best configurations reported by \newcite{Vulic2017}. We refer the interested reader to the original HyperLex paper \cite{Vulic2017} for a detailed description of the best performing baseline models.

The Supervised Directional Similarity Network (\textbf{\small{SDSN}}) achieves substantially better scores than all other tested systems, despite relying on a much simpler supervision signal.
The previous top approaches, including the Paragram+CF embeddings, make use of numerous annotations provided by WordNet or similarly rich lexical resources, while for {\small{SDSN}} and \textbf{\small{SDSN+SDF}} only use the designated relation-specific training set and corpus statistics. By also including these extra training instances (\textbf{\small{SDSN+SDF+AS}}), we can gain additional performance and push the correlation to 0.692 on the random split and 0.544 on the lexical split of HyperLex, an improvement of approximately 25\% to the standard supervised training regime.

In Table \ref{tab:examples} we provide some example output from the final {\small{SDSN+SDF+AS}} model. It is able to successfully assign a high score to (\textit{captain}, \textit{officer}) and also identify with high confidence that \textit{wing} is not a type of \textit{airplane}, even though they are semantically related. As an example of incorrect output, the model fails to assign a high score to (\textit{prince}, \textit{royalty}), possibly due to the usage patterns of these words being different in context. In contrast, it assigns an unexpectedly high score to (\textit{kid}, \textit{parent}), likely due to the high distributional similarity of these words.

\newcite{Glavas2017} proposed a related dual tensor model for the binary detection of asymmetric relations (\textit{Dual-T}).
In order to compare our system to theirs, we train  our model on HypeNet and convert the output to binary decisions. We also compare {\small{SDSN}} to the best reported models of \newcite{Shwartz2016} and \newcite{Roller:2016emnlp}, which combine distributional and pattern-based information for hypernymy detection (\textit{HypeNet-hybrid} and \textit{H-feature}, respectively).\footnote{For more detail on the baseline models, we refer the reader to the original papers.} We do not include additional WordNet and PPDB examples in these experiments, as the HypeNet data already subsumes most of them. As can be seen in Table \ref{tab:results2}, our {\small{SDSN+SDF}} model achieves the best results also on the HypeNet dataset, outperforming previous models on both data splits.


\begin{table}[t]
\centering
\small
\setlength\tabcolsep{4.5pt}
\begin{tabular}{l|ccc|ccc} \toprule
 & \multicolumn{3}{c|}{Lexical split} & \multicolumn{3}{c}{Random split}\\
 & P & R & F & P & R & F\\ \midrule
Dual-T & 70.5 & 78.5 & 74.3 & 93.3 & 82.6 & 87.6\\
HypeNet-hybrid & 80.9 & 61.7 & 70.0 & 91.3 & \textbf{89.0} & 90.1\\
H-feature & 70.0 & \textbf{96.4} & 81.1 & 92.6 & 85.0 & 88.6 \\ \midrule
SDSN & \textbf{82.8} & 84.6 & 83.7 & \textbf{94.0} & 86.7 & 90.2\\
SDSN+SDF & 82.6 & 86.0 & \textbf{84.2} & 92.8 & 88.7 & \textbf{90.7} \\ \bottomrule
\end{tabular}
\caption{Results on the HypeNet binary hypernymy detection dataset.}
\label{tab:results2}
\end{table}

\section{Conclusion} We introduce a novel neural architecture for mapping and specialising a vector space based on limited supervision.
While prior work has focused only on optimising individual word embeddings available in external resources, our model uses general-purpose embeddings and optimises a separate neural component to adapt these to the specific task, generalising to unseen data.
The system achieves new state-of-the-art results on the task of scoring graded lexical entailment. 
Future work could apply the model to other lexical relations or extend it to cover multiple relations simultaneously.

\begin{table}[t]
\centering
\small
\def\arraystretch{1.0}
\setlength\tabcolsep{6.5pt}
\begin{tabular}{cccc} \toprule
& & S & P \\ \midrule
captain & officer & 8.22 & 8.17\\
celery & food & 9.3 & 9.43\\
horn & bull & 1.12 & 0.94\\
wing & airplane & 1.03 & 0.84\\ \midrule
prince & royalty & 9.85 & 4.71\\
autumn & season & 9.77 & 3.69\\
kid & parent & 0.52 & 8.00\\
discipline & punishment & 7.7 & 3.32 \\ \bottomrule
\end{tabular}
\caption{Example word pairs from the HyperLex development set. $S$ is the human-annotated score in the HyperLex dataset. $P$ is the predicted score using the {\small{SDSN+SDF+AS}} model.}
\label{tab:examples}
\end{table}

\section*{Acknowledgments}
Daniela Gerz and Ivan Vuli\'c are supported by the ERC Consolidator Grant LEXICAL: Lexical Acquisition Across Languages (no 648909). We would like to thank the NVIDIA Corporation for the donation of the Titan GPU that was used for this research.



\bibliography{ssnhyperlex}
\bibliographystyle{acl_natbib}

\end{document}